\documentclass[10pt,twocolumn,letterpaper]{article}

\usepackage{wacv}
\usepackage{times}
\usepackage{epsfig}
\usepackage{graphicx}
\usepackage{amsmath}
\usepackage{amssymb}

\usepackage{multirow}
\usepackage[multiple]{footmisc}
\usepackage{tabularx,ragged2e,booktabs}
\usepackage{pbox}

\wacvfinalcopy % *** Uncomment this line for the final submission

\def\wacvPaperID{156} % *** Enter the wacv Paper ID here

\def\tabletext{\small}
\newcommand{\specialcell}[2][c]{%
  \begin{tabular}[#1]{@{}c@{}}#2\end{tabular}}
\begin{document}

%%%%%%%%% TITLE
\title{Compact CNN for Indexing Egocentric Videos}

% Authors at the same institution
%\author{First Author \hspace{2cm} Second Author \\
% Authors at different institutions
\author{Yair Poleg ~~~~~~~~~ Ariel Ephrat ~~~~~~~~ Shmuel Peleg\\
The Hebrew University of Jerusalem\\
Jerusalem, Israel
\and
Chetan Arora\\
IIIT\\
Delhi, India}

\maketitle
%\ifwacvfinal\thispagestyle{empty}\fi

%%%%%%%%% ABSTRACT

\begin{abstract}
While egocentric video is becoming increasingly popular, browsing it is very difficult. In this paper we present a compact 3D Convolutional Neural Network (CNN) architecture for long-term activity recognition in egocentric videos. Recognizing long-term activities enables us to temporally segment (index) long and unstructured egocentric videos. Existing methods for this task are based on hand tuned features derived from visible objects, location of hands, as well as optical flow.

Given a sparse optical flow volume as input, our CNN classifies the camera wearer's activity. We obtain  classification accuracy of $89\%$, which outperforms the current state-of-the-art by $19\%$. Additional evaluation is performed on an extended egocentric video dataset, classifying twice the amount of categories than current state-of-the-art. Furthermore, our CNN is able to recognize whether a video is egocentric or not with $99.2\%$ accuracy, up by $24\%$ from current state-of-the-art. To better understand what the network actually learns, we propose a novel visualization of CNN kernels as flow fields.

\end{abstract}

\section{Introduction}

Recent advances in wearable technologies have made the usage of head mounted camera practical. Such cameras are usually operated in `always on' mode, providing access to first person point of view which is typically not available with traditional point and shoot cameras. We refer to such videos as egocentric videos. With wearable cameras becoming increasingly affordable, egocentric video recording has become common practice in many areas such as sports, hiking, and law enforcement. While the possibility of sharing one's actions with the community is compelling enough, usage of such cameras for life logging is also on the rise.

\begin{figure}[t]
    \centering
	\includegraphics[width=1\linewidth]{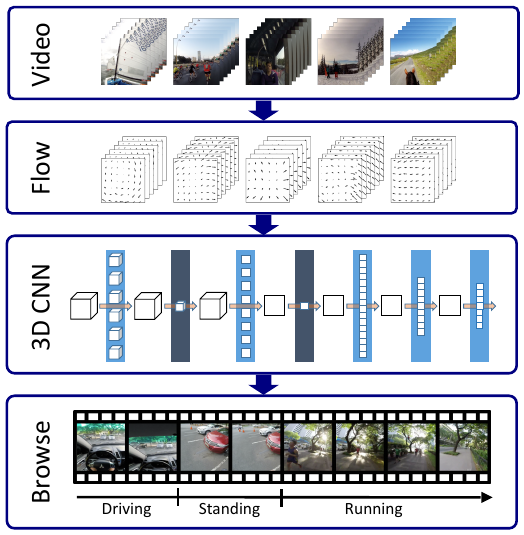}
    \caption{The proposed pipeline for classifying video segments in egocentric video according to the activity of the camera wearer. The CNN takes as input sparse optical flow from several frames arranged in a 3D array. 3D convolutions and 3D pooling are used in the network to preserve temporal structure and facilitate the learning of long term features.}
    \label{fig:overview}
\end{figure}

Egocentric video gives a novel perspective to capture social and object interactions, but also poses new challenges for the computer vision community. The endless motion from head mounted cameras and the unstructured nature of videos that are shot in an 'always on' mode make egocentric videos challenging to analyze. Recent papers in the field explored a variety of topics such as object recognition \cite{fathi_ego_objects, ego_handled_objects, ego_obj2}, activity recognition \cite{fathi_daily_act, fathi_action_from_gaze, jpl, ego_act_temporal1, ego_adl, ego_ac_recog_5_lowres, ego_ac_recog_eyecam, ego_ac_recog_cvprw14}, summarization \cite{grauman-important-people, grauman-story, ego_novelty}, gaze detection \cite{fathi_predict_gaze} and social interactions \cite{ego_social}. Other tasks such as temporal segmentation \cite{us, kitani}, frame sampling \cite{grauman-snap-points,egosampling}, hyperlapse \cite{hyperlapse} and camera wearer identification \cite{us_accv14,ego_biometrics,egosurfing} have been explored as well.

\begin{figure*}[!th]
    \centering
	\includegraphics[width=1\linewidth]{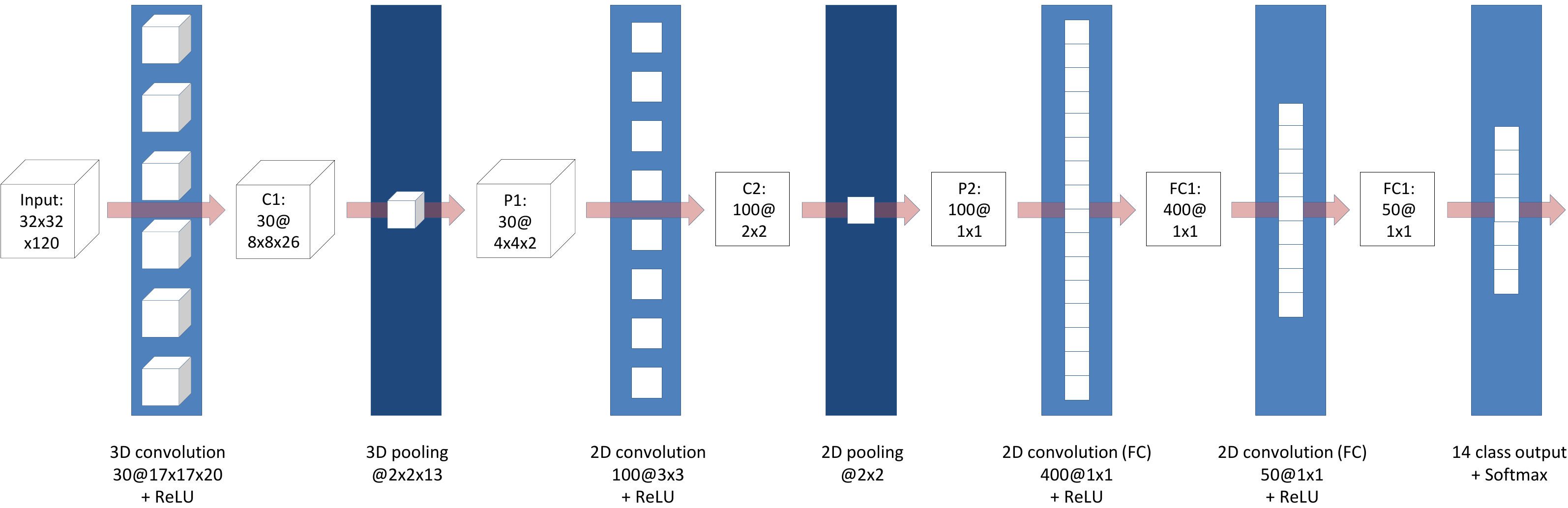}
    \caption{Network architecture. Our network takes as input sparse optical flow computed on a grid of $32 {\times} 32$, stacked over $60$ frames. The first hidden layer is a rather long $3D$ convolution followed by $3D$ pooling. Long $3D$ convolutions allow the network to learn features for long term activities, the focus of this paper. Next layers are standard $2D$ convolutional and pooling layers, followed by $2$ fully connected layers and terminating with a softmax layer. All hidden weight layers use the ReLU activation function.}
    \label{fig:netarch}
\end{figure*}

Encouraged by the success of neural networks, we investigate CNN architecture for determining the activity of the camera wearer in egocentric videos. Obtaining large enough datasets to train CNNs is always a challenge. In the case of egocentric vision, it is even more difficult due to legal and ethical issues  (e.g. privacy). To overcome the data scarcity problem, we propose a compact CNN which is trained on optical flow instead of pixel intensities.
Using motion cues for egocentric vision tasks aligns well with recent works in the field (e.g. \cite{jpl, kitani, ryoo_pooling, us, us_accv14,ego_or_not_cvprw14}).  Fig.~\ref{fig:overview} illustrates the proposed pipeline for classifying egocentric videos according to the activity of the camera wearer.

To explore the capabilities of the proposed architecture, we compare it to Poleg \etal \cite{us}, who used sparse optical flow (without appearance based features) for classifying video into $7$ long term activities such as walking, driving, etc. We obtain $89\%$ accurate classification, $19\%$ higher than their original method. We also extend their dataset from 7 to  14 activity classes, and achieve an accuracy of $86\%$ on this new dataset. We also test our network on the task of determining whether a video is egocentric or not. We report accuracy of $99\%$, up by over $27\%$ on the classification accuracy reported by Tan \etal \cite{ego_or_not_cvprw14}. They suggested to use approximately 50 handcrafted features for this task, based on both intensities and optical flow.

In an attempt to understand what the network actually learns, we analyze the kernels learned by our network for the task of long-term activity recognition. Previous works \cite{twostream, 3dconv_action_pami13, cnn_action_tubes, rcnn,modeep,cnn_visu} visualize learned kernels by mapping kernel weights to image intensities. We find this style of visualization difficult for analyzing optical flow based kernels. Instead, we suggest visualizing the learned kernels as optical flow fields. Using this visualization, we show that the proposed CNN learns intuitive and meaningful kernels.

The rest of this paper is organized as follows. We survey related works in Section \ref{sec:related_work}. We present the proposed CNN architecture in Section \ref{sec:cnn_arch}. In Section \ref{sec:experiments} we evaluate the power of our network on several egocentric tasks and provide insights by analyzing the learned convolutional kernels. We conclude in Section \ref{sec:concl}.

\section{Related Work}
\label{sec:related_work}

\paragraph*{Activity Recognition in Egocentric Video}

There are two main approaches to activity recognition in egocentric video. The first approach is based mainly on appearance, in which object detection, hand recognition and other visual cues are used in order to infer the activity performed by the camera wearer \cite{fathi_daily_act, fathi_action_from_gaze,ego_adl,ego_ac_recog_cvprw14,ego_ac_recog_5_lowres}. These methods perform best for short-term tasks in a controlled environment. In this work we focus on long-term activity performed by the camera wearer in the wild.

The second line of works for activity recognition is based on motion analysis. Ogaki \etal \cite{ego_ac_recog_eyecam} used motion cues from both an eye-tracking camera and a head-mounted camera to infer egocentric activities. Spriggs \etal \cite{ego_act_temporal1} used inertial motion sensors coupled with a head mounted camera. Our work is based  only on a standard head mounted camera. Kitani \etal \cite{kitani} used frequency and motion based features to learn short-term actions in an unsupervised setting. Ryoo and Matthies \cite{jpl} used global and local motion features to recognize interaction-level activities. The goal of our work is to recognize a set of pre-defined, semantically meaningful, long-term activities.

Recently, Ryoo \etal \cite{ryoo_pooling} proposed a new feature representation framework based on time series pooling, which is able to abstract detailed short-term/long-term changes in motion/appearance descriptor values over time. In \cite{us} temporal filtering is applied to sparse optical flow in order to recognize long-term activities.  Our work is inspired by these works. We generalize the concept of temporal filtering by learning a set of long-term 3D convolution kernels. We apply 3D pooling operators to add robustness against temporal shifts and speed variations in the activities. We report an improvement of $19\%$ on the long term activity recognition task of \cite{us} using our compact CNN architecture.

\paragraph*{CNNs for Video using Intensities}

In recent years several papers have used CNN for video analysis problems. Karpathy \etal suggest CNNs for large scale (non egocentric) video classification using a dataset of 1 million YouTube videos, spanning 487 classes \cite{sports1m}. We note the authors' remark that in their architecture there is only a minor difference in the results when  operating on single video frame or on a stack of frames. This might indicate that their learned spatio-temporal features do not capture motion well. The fact that motion plays a major role in egocentric vision and the sheer amount of data required to train their network makes their approach less practical for our goals.

Tran \etal propose to learn generic features for video analysis by training a deep 3D convolutional neural network \cite{c3d}. They show that by using $3D$ (spatio-temporal) convolutions, the learned features encapsulate appearance and motion cues. The power of their features is demonstrated on several tasks, including an object recognition task from egocentric videos. In this work we focus on long-term activity recognition from egocentric videos.

\paragraph*{CNNs for Video using Optical Flow}

A compact CNN using sparse optical flow to identify the camera wearer from egocentric video was reported in \cite{ego_biometrics}. Jain \etal \cite{modeep} use a combination of image intensities and instantaneous optical flow from a pair of frames for human pose estimation. Gkioxari and Malik \cite{cnn_action_tubes} also use instantaneous optical flow and image intensities to perform action recognition and localization. Ji \etal \cite{3dconv_action_pami13} use image intensities and optical flow of 5 consecutive frames for human action recognition. To capture the motion component effectively, Simonyan and Zisserman \cite{twostream} propose a two stream model where one stream  handles a stack of image intensities and the other handles a stack of frame-to-frame dense optical flow fields.

Our work differs from all the aforementioned works in several ways: (i) We suggest a CNN architecture tailored for the unique challenges of activity recognition in egocentric video. (ii) Our network is compact with respect to the previous CNNs for activity recognition, enabling training on the relatively small datasets currently available to the egocentric vision community.  (iii) The input to our network is built from sparse optical flow (a fixed grid of only $32 {\times} 32$ flow vectors) while all previous CNNs for activity recognition used dense optical flow.

\section{CNN Using Sparse Optical Flow}
\label{sec:cnn_arch}

\subsection{Network Input}

\begin{figure}[t]
    \centering
	\includegraphics[width=0.9\linewidth]{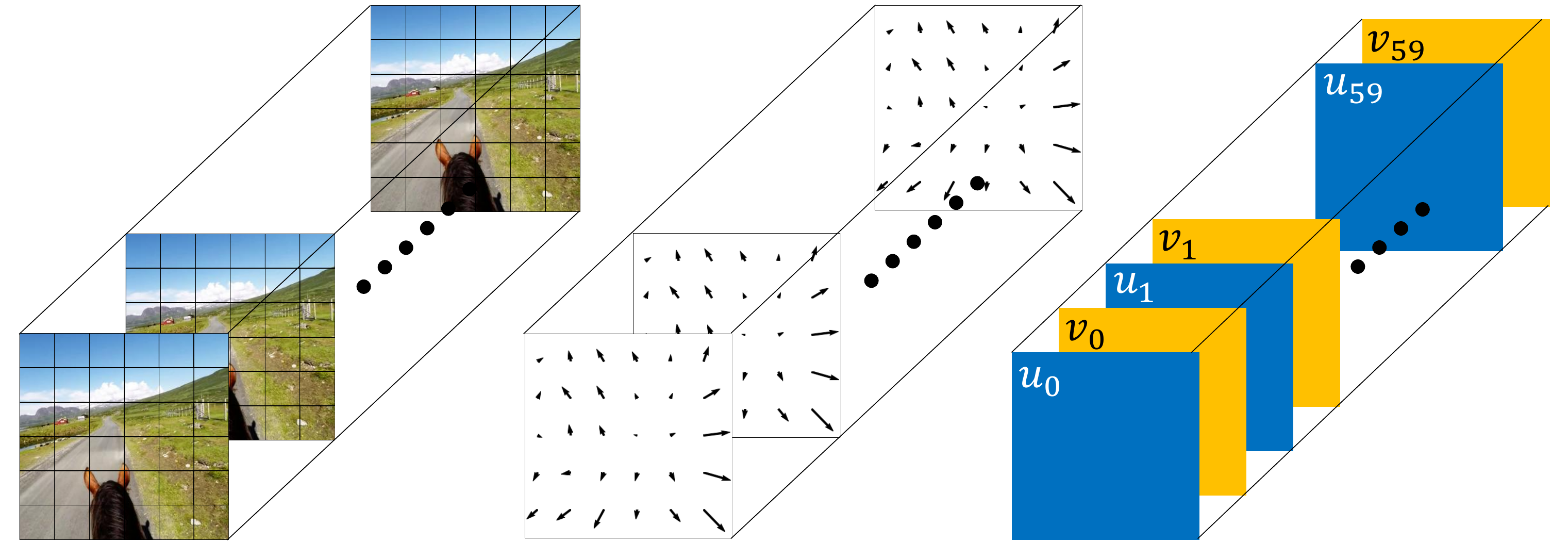}
    \caption{The input to our network is sparse optical flow. Left and Center: We divide each frame into a grid of $32 \times 32$ cells and find optical flow independently between each corresponding grid cells of two consecutive frames (one global $(x,y)$ translation pair for each grid cell). Right: We alternately stack the $x$ and $y$ optical flow components corresponding to $60$ consecutive frames. The network's input is a data volume of $32 {\times} 32 {\times} 120$ elements. Stacking optical flow from such a large number of frames enables the network to learn features for long term activities.}
    \label{fig:input-explanation}
\end{figure}

\begin{figure*}[t]
    \centering
	\includegraphics[width=1\linewidth]{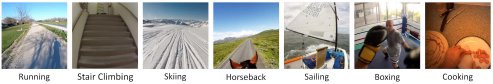} \\
    \includegraphics[width=1\linewidth]{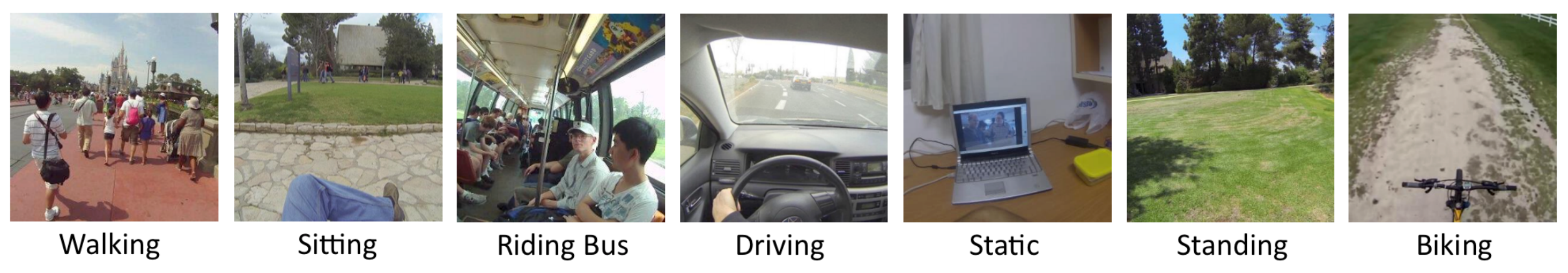}
    \caption{We extend the dataset of \cite{us} with an additional $16$ hours of video, introducing $7$ new activity classes. This figure shows representative frames from the $7$ new activity classes we introduce (top row) as well as from the original $7$ classes. All together, we have $14$ activity classes. See Section \ref{sec:experiments} for details.}
    \label{fig:thumb_7_new_classes}
\end{figure*}

First person activity in egocentric videos is usually manifested over multiple frames. Previous works in egocentric activity recognition have therefore derived features from $2$ seconds of video for short term actions \cite{kitani}, and up to $17$ seconds of video for long term activities \cite{us}. The input to our network is sparse optical flow derived from $4$ seconds. We note the observation made in \cite{twostream} that their network performs better when it doesn't need to learn to estimate motion implicitly.

Given a video sequence, we begin by normalizing its frame rate to $15$ FPS (frames per second). Next, we divide each video into overlapping blocks of $4$ seconds ($60$ frames). The overlap between consecutive blocks is $2$ seconds ($30$ frames). We then compute sparse optical flow vectors as described in \cite{us,us_accv14,ego_biometrics} by dividing each frame into a non-overlapping grid of size $32 {\times} 32$. Optical flow is then estimated between two corresponding grid cells in consecutive frames using LK \cite{lk}. An optical flow field is represented by a $32 {\times} 32 {\times} 2$ volume, corresponding to the flow in $x$ and $y$ directions. The input to our network is created by stacking $60$ such flow fields, resulting in a volume of  $32 {\times} 32 {\times} 120$ scalars (see Fig. \ref{fig:input-explanation}).

Formally, let $u_k(i,j)$ and $v_k(i,j)$ denote optical flow in $x$ and $y$ directions between grid cells $(i,j)$ of frame $k$ and frame $k+1$. Let $t$ denote the time instance (frame number) for which we want to assign an activity label. The input to our CNN at time instance $t$ is the volume $I_{t} \in \mathbb{R}^{32 {\times} 32 {\times} 120}$, which is computed from the optical flow as follows:
\begin{equation}
\begin{split}
&I_{t}(i,j,2\tau) = u_{t+\tau}(i,j)\\
&I_{t}(i,j,2\tau+1) = v_{t+\tau}(i,j)  \label{eq:input-1}
\end{split}
\end{equation}
where $\tau$ is in the range of $0..59$.

Data normalization for CNNs has become standard practice. We perform the following normalization procedure. Once we obtain all the input volumes $I_{\tau}$ of a certain video dataset, we find the $95$th percentile of $u_t(i,j)$ and $v_t(i,j)$ separately and clamp all values to the respective $95$th percentile value. We then scale the data to the range $[-1,1]$.

\subsection{Network Architecture}

The first hidden layer in our network C1 is a $3D$ convolutional layer. In this layer we learn $30$ kernels of size $17 {\times} 17 {\times} 20$. The kernels are applied with a spatial stride of $2$ and temporal stride of $4$ (a stride along the 3rd dimension). By keeping the temporal stride even we ensure that the learned kernels do not mix optical flow from $x$ and $y$ directions. We apply the Rectified Linear Unit (ReLU) non-linearity to the output of this layer.

The feature maps generated by each of the $30$ kernels of C1 are of size  $8 {\times} 8 {\times} 26$. All these outputs are concatenated along the 3rd dimension to give an output volume of size $8 {\times} 8 {\times} 780$. The exact start time and duration of an activity within an input block can vary. To overcome this, layer P1 pools the feature volume generated by C1 using a 3D max pooling operator of size $2 {\times} 2 {\times} 13$ and a temporal stride of $13$. This means that each feature map generated by a single kernel in C1 is pooled temporally exactly twice. Therefore, the output of P1 is a volume of size $4 {\times} 4 {\times} 60$.

The second hidden layer C2 is a standard $2D$ convolutional layer with $100$ kernels of size $3 {\times} 3$, followed by a ReLU and a max pooling operator of size $2 {\times} 2$.
Layers FC1 and FC2 are fully connected of size $400$ and $50$ respectively. The last layer is a softmax with the number of nodes equal to the number of classes at hand (application dependent). In total, our network contains $287,400$ trainable variables, two orders of magnitude less than \cite{twostream,c3d}.  Fig.~\ref{fig:netarch} shows the complete network structure.

\subsection{Classification with Temporal Context}
\label{classification_with_temporal_context}
The duration of long-term activities can range from a few seconds to minutes. The input to our CNN, on the other hand, covers only $4$ seconds. Contradicting short-term actions (e.g. pausing for a second while running) might result in sporadic misclassifications (outliers). As in \cite{ego_biometrics}, we add up the softmax scores of $\eta$ consecutive samples and select the activity label with the highest score. This reflects our prior domain knowledge (e.g. skiing for $4$ seconds while running is unlikely). We have explored several values for $\eta$ in the range of $[0,30]$ and found that $\eta=21$, which is equivalent to $44$ seconds, gives good results.

\subsection{Design Choices}
In deep learning, the objective's dependency on hyperparameters is notoriously complex. This problem is amplified when the dataset has its own set of hyperparameters, as in our case (e.g. flow field resolution, temporal receptive field, output filtering etc.).
Our primary goal was to design a CNN that performs well on the egocentric tasks at hand, while keeping the number of trainable variables to a minimum to overcome the data scarcity problem.

We first used the $10 {\times} 5$ flow field as in \cite{us}. Temporal receptive fields of $32,60,120,$ and $240$ frames did not yield satisfactory results. We believe this is because the CNN needs more spatial information for the convolution to effectively learn recurring spatial motion patterns in the data. The large 3D kernel size which we converged to ($17 {\times} 17 {\times} 20$) seems to support this reasoning by indeed learning ``meaningful'' spatio-temporal patterns.

Increasing the spatial resolution of the flow field to $32 {\times} 32$ and performing a similar temporal length search fared much better. It is possible that a denser optical flow could have performed even better, but with a drastic increase in size of input data and trainable parameters. Reducing the input temporal length to 4 second blocks also helps reduce the number of trainable parameters. Activity duration is accounted for during post-processing, using the scheme described in Sec. \ref{classification_with_temporal_context}.

As with CNNs, a better combination of data and network hyperparameters may exist. 

\subsection{Visualization of Learned Kernels}
\label{subsec:visualization}

\begin{figure}[t]
    \centering
	\includegraphics[width=0.6\linewidth]{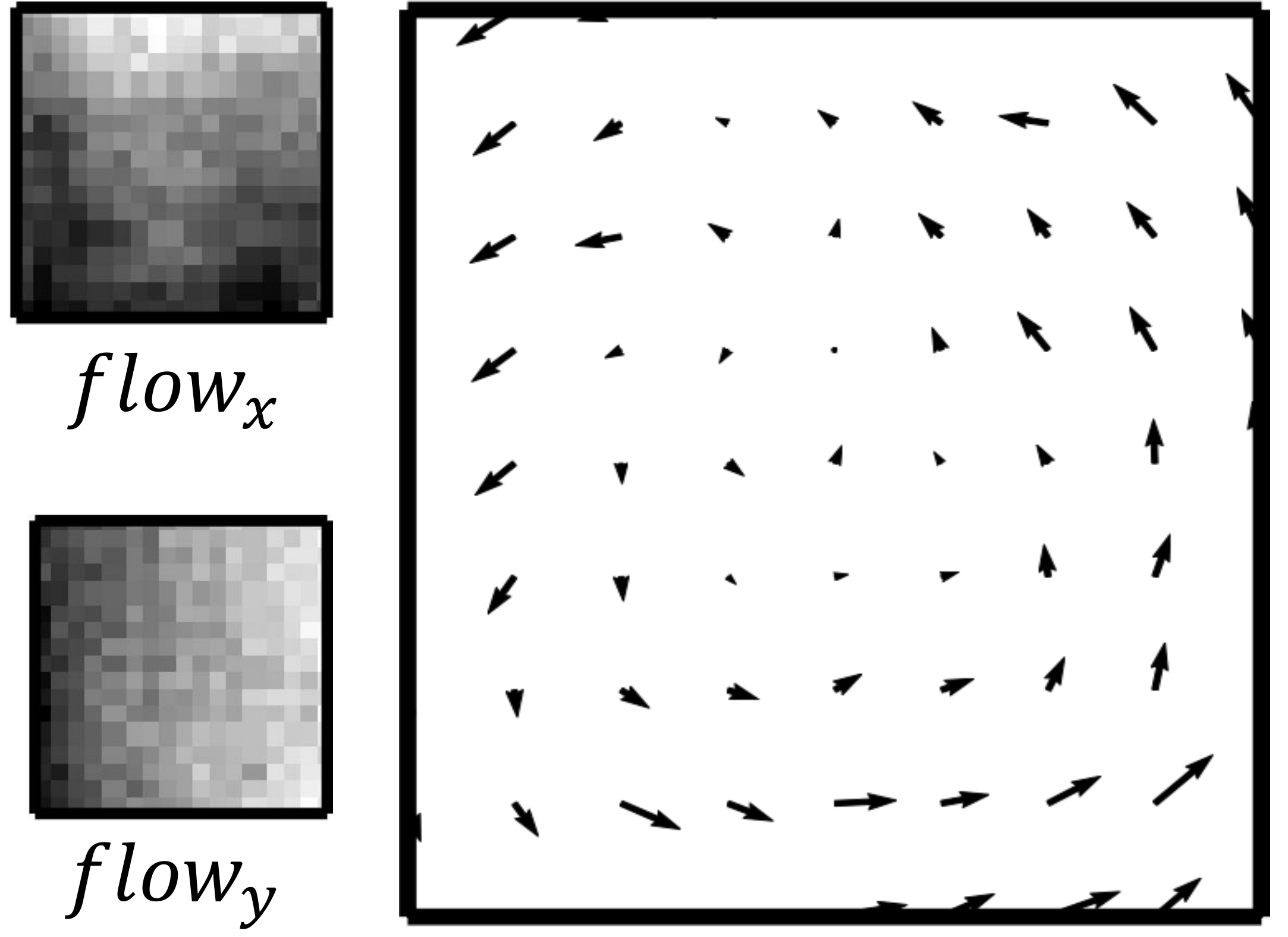}

    \caption{A slice from a $17 {\times} 17 {\times} 20$ kernel learned by our network for the task of temporal segmentation. The kernel has strong affinity to the \emph{Walking} class. Note: For clarity we show only half of the flow vectors. Visualization of learned kernels by pixel intensities is less informative for a network using optical flow as input. We suggest a visualization where the weights of a $3D$ convolution kernel from two adjacent slices are displayed as an optical flow field. The two images on the left show weights as intensities. The figure on the right shows the same weights as an optical flow field. With this visualization it is clear that the motion pattern learned by the kernel is rotation. }
    \label{fig:kernel_visu}
\end{figure}

One way of gaining insight into a fully-trained CNN is through visualization of its learned kernels and input activations  \cite{cnn_visu}. When natural images are used as network input, kernel weights are usually mapped to intensities and visualized as an image. However, the input to our network is a volume composed of stacked flow fields. We therefore suggest visualizing the kernels as vector fields instead of intensity images. We note that each kernel is composed of slices that represent optical flow in the $x$ and $y$ directions. Each such pair is then visualized as flow field. This essentially inverts the stacking process of Equation \ref{eq:input-1}.  Fig.~\ref{fig:kernel_visu} shows an example in which the learned z-rotation motion pattern is hard to infer from the intensity images, but becomes clear when visualized as a vector field. While this may seem trivial, to the best of our knowledge it has not done before in the field of CNNs \cite{ego_biometrics,modeep,cnn_action_tubes,3dconv_action_pami13,twostream}.

\section{Experiments}
\label{sec:experiments}

\begin{figure*}[t]
    \centering
	\includegraphics[width=1\linewidth]{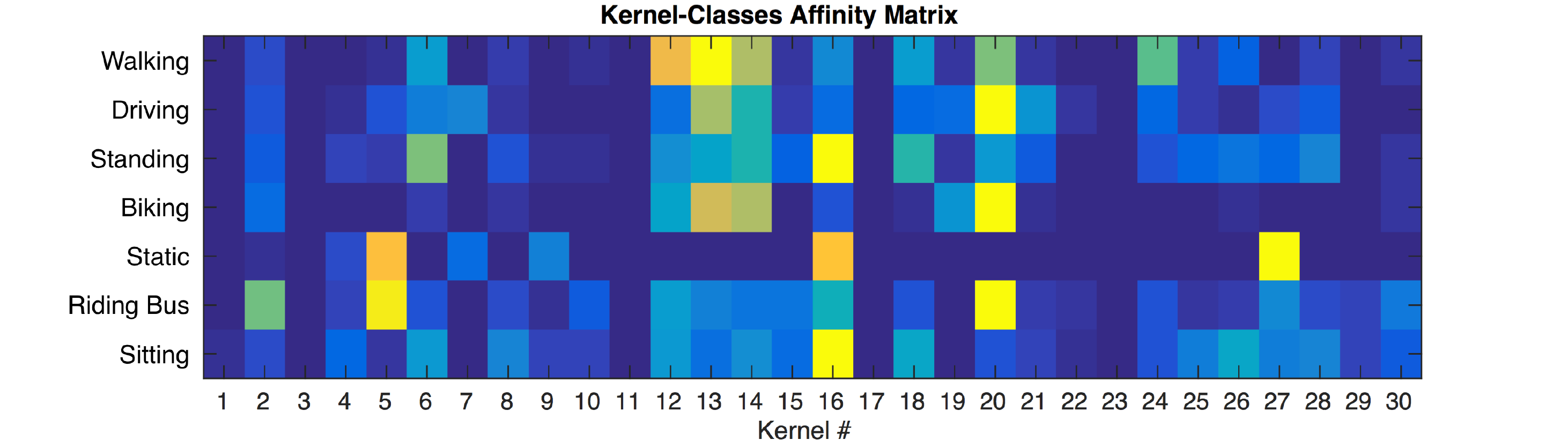}
    \caption{Class-kernel affinity matrix for the first hidden layer. For each sample of a particular class, we find the kernel giving the strongest response. The class-kernel affinity is defined as the number of times the kernel has the strongest response for samples of this class. Brighter colors mean stronger affinity (more votes).
    We can see in this matrix that kernel $\#16$ has strong affinity to \emph{Standing, Static} and \emph{Sitting}, whereas kernel $\#20$ seems to be popular for \emph{Riding Bus, Biking} and \emph{Driving}. These classes are indeed semantically closer to each other.
    This indicates that the proposed network may be able to learn meaningful kernels. We cautiously speculate that such affinity matrices can be used to design class hierarchies for classification.}
    \label{fig:kernel_classes_affinity}
\end{figure*}

In this section we evaluate the performance of the proposed network for several activity recognition tasks in egocentric vision. We do not fine-tune the parameters our network for the different experiments and keep the architecture fixed throughout the experiments.

\paragraph{Implementation Details} Our network implementation is based on Caffe \cite{caffe}. The parameters for the network are fixed as follows: Network weights are initialized using the \emph{Xavier} normalized initialization procedure \cite{xavier}. The learning rate for all convolutional and fully connected layers is set to $0.01$. We use mini-batches of $64$ training samples each and stop training after $3000$ iterations. It takes $15$ minutes to train a network to classify $14$ classes with $1300$ training samples per class using a single \emph{Nvidia Titan Black} GPU. 

Sparse optical flow is estimated using the implementation released by \cite{us}. Their implementation is based on LK \cite{lk}. In the rare cases that LK fails to converge, we interpolate the flow value from temporally adjacent frames.

\subsection{Temporal Segmentation into 7 Classes}
\label{egoseg_exp}

\paragraph{Dataset:} We evaluate the performance of our architecture on the dataset of \cite{us}. Their dataset contains $65$ hours of egocentric videos collected from multiple subjects at Disney World \cite{ego_social}, YouTube and others. The subjects perform various tasks, both indoors and outdoors. The dataset contains annotation for $7$ activity classes: \emph{Walking, Driving, Riding Bus, Biking, Standing, Sitting, Static}. Two classes out of the $7$ (\emph{Static} and \emph{Bus}) had less than $30$ minutes of video, which reduced to less than $900$ samples per class. In order to train our network, we shot additional sequences for these categories using a \emph{GoPro Hero3} camera. The sequences we shot are similar to the original sequences in the dataset in terms of content (scenario), resolution, FPS and even the camera make. We will release the new sequences and their annotations.

\renewcommand{\tabcolsep}{0.2cm}
\begin{table}
    \centering
    \begin{tabular}{lccccccc}
        \toprule[1.5pt]
        & \multicolumn{4}{c}{\bf Recall} \\
        \specialcell{\bf Class} & \specialcell{\bf \cite{us}} & \specialcell{\bf \cite{ryoo_pooling}} & \specialcell{\bf \cite{c3d}} & \specialcell{\bf Ours} \\ \midrule
         
Walking  &  $83\%$  &  $\bf 91\%$  &  $79\%$       &  $89\%$          \\
Driving      &  $74\%$  &  $82\%$      &  $92\%$       &  $\bf 100\%$     \\
Standing &  $47\%$  &  $44\%$      &  $62\%$       &  $\bf 79\%$      \\
Riding Bus      &  $43\%$  &  $37\%$      &  $58\%$       &  $\bf 82\%$      \\
Biking   &  $86\%$  &  $34\%$      &  $36\%$       &  $\bf 91\%$      \\
Sitting  &  $62\%$  &  $70\%$      &  $62\%$       &  $\bf 84\%$      \\
Static   &  $97\%$  &  $61\%$      &  $\bf 100\%$  &  $98\%$          \\
        \midrule
\bf Mean     &  $70\%$  &  $60\%$      &  $70\%$       &  $\bf 89\%$      \\
        \bottomrule[1.5pt] \\
    \end{tabular}
    \caption{Our proposed method does $19\%$ better compared to \cite{us} and \cite{c3d} on $7$ class activity recognition, and $29\%$ better compared to \cite{ryoo_pooling}. Our method achieves a high accuracy on \emph{Driving}, while \emph{Sitting, Standing} and \emph{Riding Bus} continue to be the more difficult classes to recognize, as in \cite{us}. }
    \label{tb:egoseg_comparison}
\end{table}

\paragraph{Results:} We follow evaluation protocol of \cite{us} for this task. We randomly pick sequences until we have $1300$ samples (approximately $90$ minutes of video) per class. A sequence from which we take samples for the training set is disqualified from the test set. Table \ref{tb:egoseg_comparison} details the classification results. The proposed network is able to reduce the error in all the classes with respect to \cite{us}. The mean error is reduced from $30\%$ to $11\%$. In addition, we used the code released by \cite{ryoo_pooling} and \cite{c3d} to compare with their methods. Our method outperforms theirs by $29\%$ and $19\%$, respectively. The worst accuracy is observed for class \emph{Standing}, similar to the current state-of-the-art.

\paragraph{Analysis:} To gain insight into what the network is learning, we analyze the kernels of the first convolutional layer (C1). We run the test set through the network again and for each sample find the top $3$ kernels that give the highest response with that sample. Each sample then `casts a vote' to each of these top $3$ kernels. The votes are accumulated in an affinity matrix. We expect kernels that received significantly more votes from a particular class to have a strong affinity to that class. Fig.~\ref{fig:kernel_classes_affinity} visualizes the affinities between kernel and classes for the $7$-class classification task at hand. Brighter colors mean stronger affinity (more votes). We investigate dominant kernels from each class by visualizing their weights as flow fields. Fig.~\ref{fig:kernel_visu} shows the last slice of kernel \#$24$, which has strong affinity with \emph{Walking}. This slice captures rotation about the z-axis, which is a common instantaneous motion in egocentric video captured while walking. We show additional intuitive examples in the next experiment (See Sec. \ref{sec:egoseg_plus}). It is interesting to see that certain kernels are dominant in several classes. For example, kernel $\#16$ has strong affinities to \emph{Standing, Static} and \emph{Sitting}. Similarly, kernel $\#20$ has strong affinities to \emph{Riding Bus, Biking} and \emph{Driving}. We cautiously regard this observation as another indication that kernels at the first convolutional layer capture characteristic motion patterns.

\begin{figure}[t]
    \centering
	\includegraphics[width=0.6\linewidth]{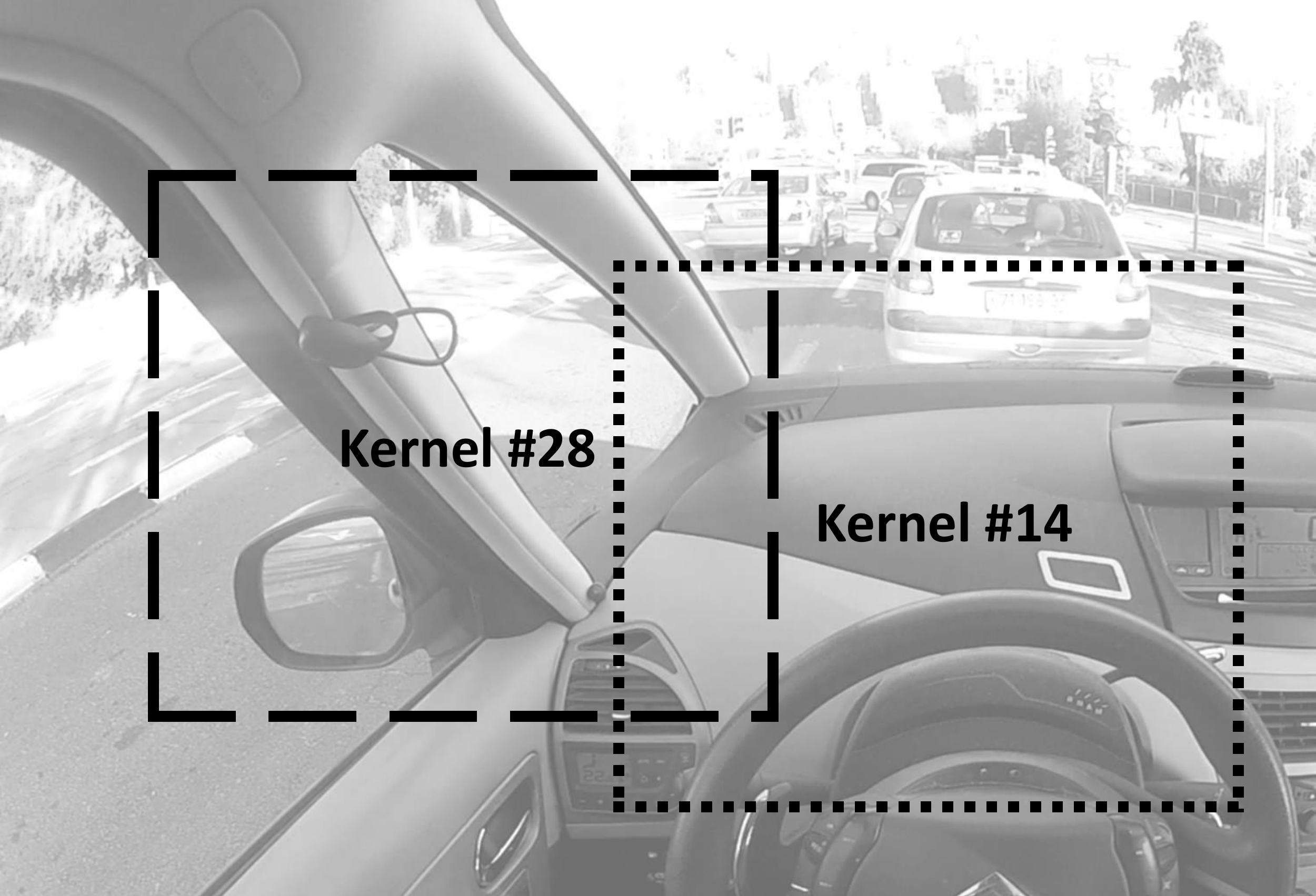} \\	\includegraphics[width=0.6\linewidth]{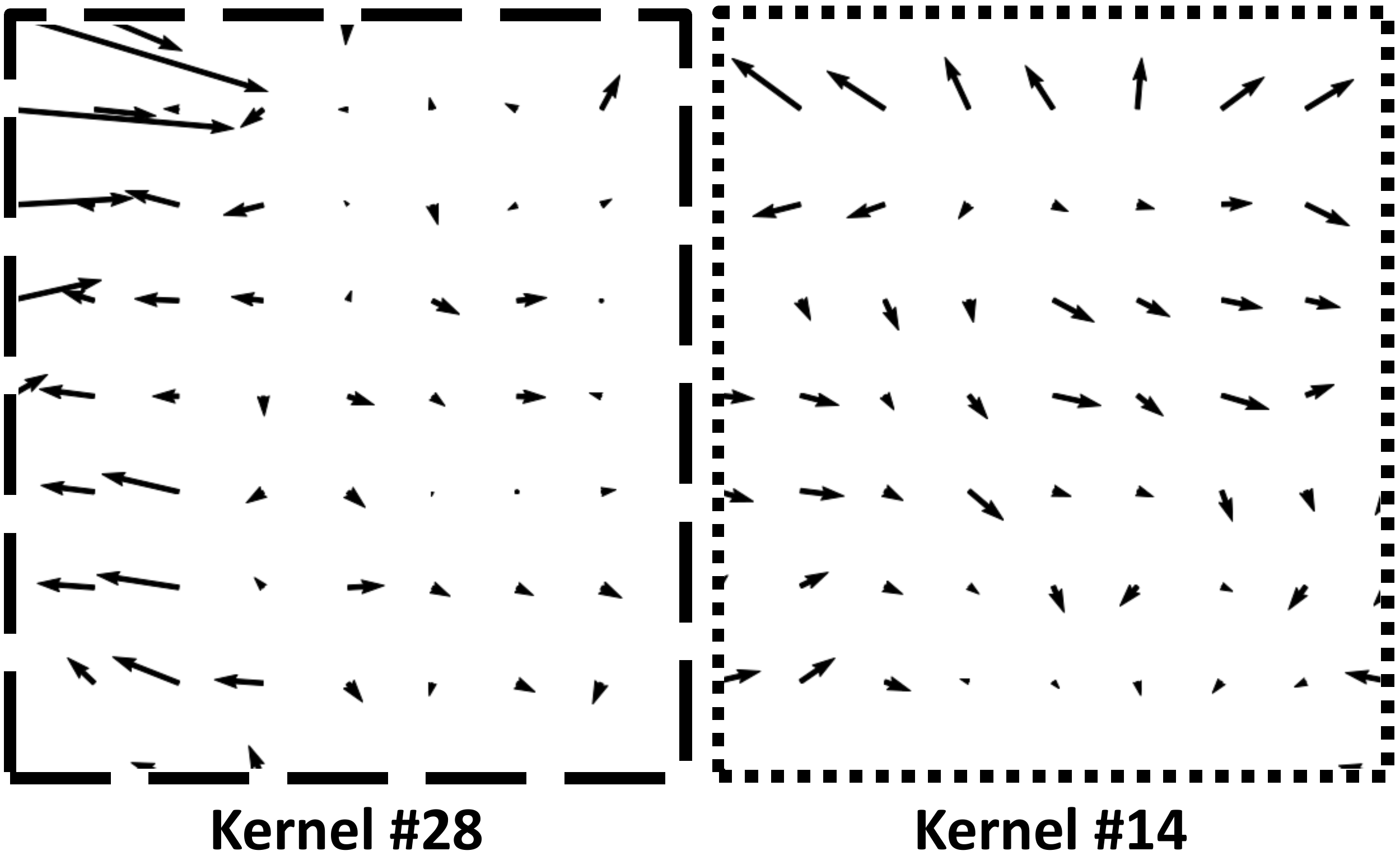}
    \caption{Two slices from two kernels having strong affinity to the \emph{Driving} class (bottom) and possible matching locations (top). The kernels learn `mixed' flow patterns, that represent locations inside the car and through the car's windows.  }
    \label{fig:kernel_driving}
\end{figure}

\begin{figure*}[t]
    \centering
	\includegraphics[width=0.75\linewidth]{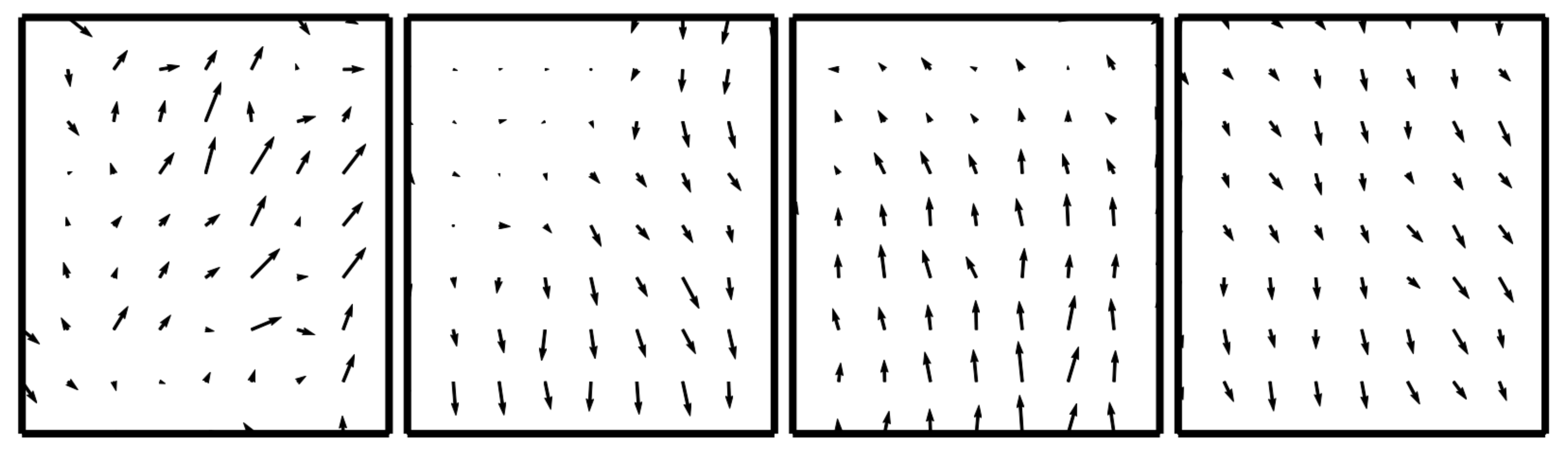}
    \caption{The first 4 slices of the kernel having the strongest affinity for the \emph{Running} class. The flow field visualization helps to see that the kernel learns alternating vertical motions (up/down). It seems that the kernel captures the sharp camera vibration that occurs when the runner's foot hits the ground.}
    \label{fig:kernel_running}
\end{figure*}

\subsection{Temporal Segmentation into 14 Classes}
\label{sec:egoseg_plus}

\paragraph{Dataset:} We extended the original $7$-class dataset of \cite{us}. Data for $6$ additional classes ($14$ hours of YouTube videos) was collected by Mechanical Turkers. The \emph{GTEA Gaze+} dataset \cite{fathi_action_from_gaze} is used in its entirety to form a $7$th new class (\emph{Cooking}). The new classes are: \emph{Running, Stair Climbing, Skiing, Horseback Riding, Sailing, Boxing} and \emph{Cooking}. Fig. \ref{fig:thumb_7_new_classes} shows a representative frame from each new class. The data is distributed evenly across the $7$ new classes with roughly $2$ hours of footage per class. We have manually annotated the new dataset and will release it and its annotations. All together we have about $82$ hours of annotated data for $14$ activity classes. Here too we use $1300$ training samples (approximately $90$ minutes of video) per class.
\\

\begin{figure}[t]
    \centering
	\includegraphics[width=1\linewidth]{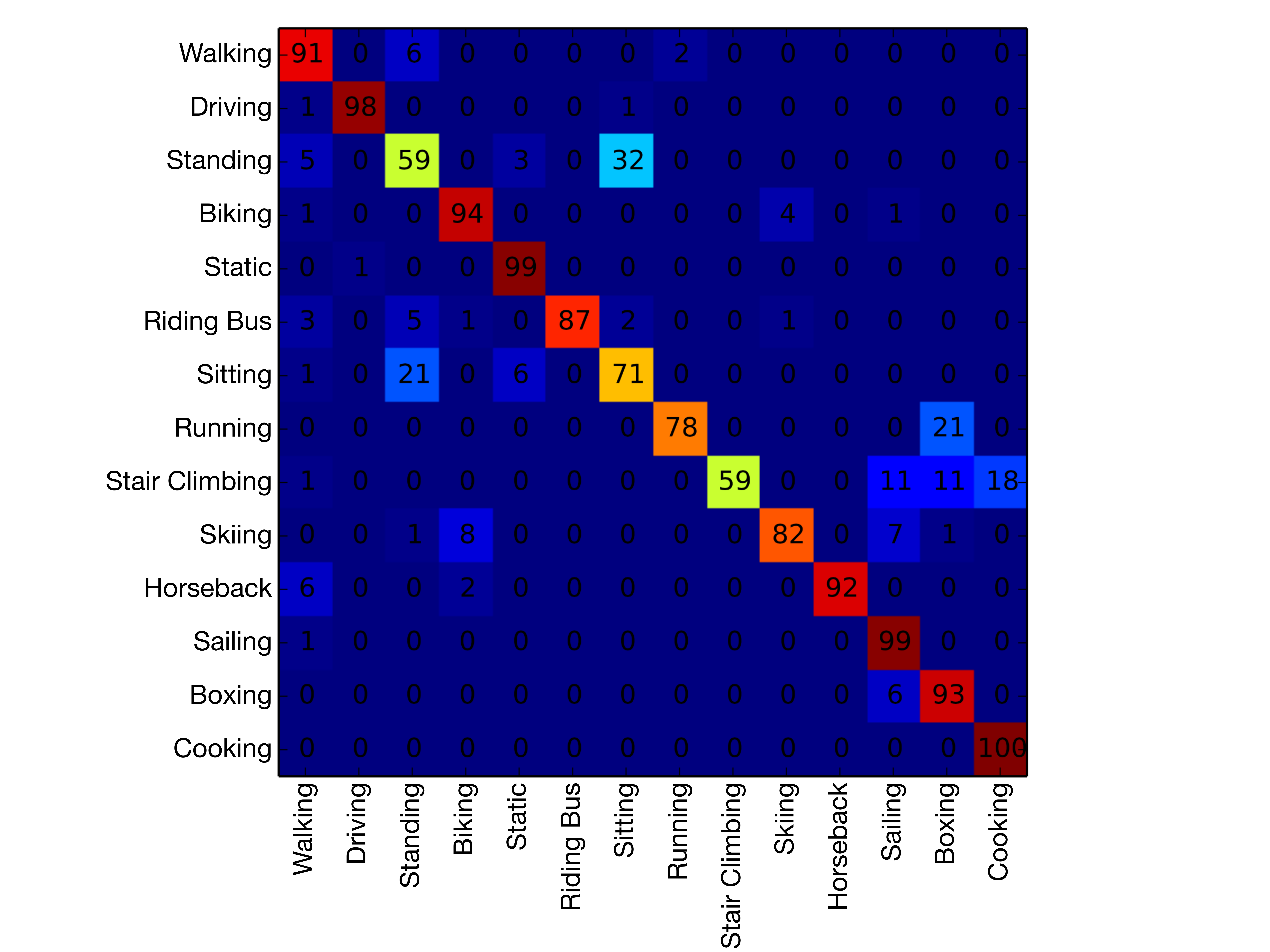}
    \caption{Confusion matrix for 14 activity classes recognition task. A confusion between \emph{Sitting} and \emph{Standing} is evident, similar to \cite{us}. There is room for improving the network's ability to distinguish between \emph{Stair Climbing} and \emph{Sailing}, \emph{Boxing} and \emph{Cooking}.}
    \label{fig:confmat_14_classes}
\end{figure}

\paragraph{Results:} We evaluate the performance in terms of Precision, Recall and F-score. The F-score is the harmonic mean of the precision $P$ and recall $R$, defined as $F = 2PR/(P+R)$. Values range from $0$ to $1$, where $1$ represents perfect performance. Table \ref{tb:prec_recall_14_classes} gives the results of our experiments on the $14$ activity classes. We achieve an average recall rate of $86\%$ for this task, a drop of $3\%$ with respect to our performance on the original $7$ classes only. This performance drop is expected since we doubled the amount of classes. We note that the network is able to achieve a recall rate of $100\%$ for \emph{Cooking} and $99\%$ for \emph{Sailing} as well as \emph{Static}.  Fig. \ref{fig:confmat_14_classes} shows the full confusion matrix for this task.

\renewcommand{\tabcolsep}{0.1cm}
\begin{table}
    \centering
    \begin{tabular}{lccccccc}
        \toprule[1.5pt]
        \specialcell{\bf \tabletext Class}  &\specialcell{\bf \tabletext Precision } & \specialcell{\bf \tabletext Recall} &\specialcell{\bf \tabletext F1-Score} \\  \midrule

Walking    &  $0.93$   &   $0.91$   &   $0.92$ \\
Driving    	   &  $0.94$   &   $0.98$   &   $0.96$ \\
Standing   &  $0.62$   &   $0.59$   &   $0.60$ \\
Biking     &  $0.92$   &   $0.94$   &   $0.93$ \\
Static     &  $0.44$   &   $0.99$   &   $0.61$ \\
Riding Bus &  $0.94$   &   $0.87$   &   $0.91$ \\
Sitting    &  $0.73$   &   $0.71$   &   $0.72$ \\
Running    &  $0.91$   &   $0.78$   &   $0.84$ \\
Stair Climbing &  $1.00$   &   $0.59$   &   $0.74$ \\
Skiing     &  $0.92$   &   $0.82$   &   $0.87$ \\
Horseback  &  $1.00$   &   $0.92$   &   $0.96$ \\
Sailing    &  $0.65$   &   $0.99$   &   $0.79$ \\
Boxing	   &  $0.47$   &   $0.93$   &   $0.62$ \\
Cooking	   &  $0.80$   &   $1.00$   &   $0.89$ \\

		\midrule
\bf Mean	   & $\bf 0.80$ & $\bf 0.86$ & $\bf 0.81$ \\
        \bottomrule[1.5pt] \\
    \end{tabular}
    \caption{Our network achieves an average recall rate of $86\%$ in the $14$ activity classes task. The classes \emph{Drving, Static, Sailing and Cooking} achieve near perfect recall rates. On the other hand, the network struggles with the \emph{Stair Climbing} and \emph{Standing} classes. See confusion matrix in Fig.~\ref{fig:confmat_14_classes}.}
    \label{tb:prec_recall_14_classes}
\end{table}

\paragraph{Analysis:}
Similar to the analysis done in the previous experiment (Sec. \ref{egoseg_exp}), for each class we find the C1 kernels with the strongest affinity. Fig. \ref{fig:kernel_running} shows $4$ slices from the kernel that is most associated with \emph{Running}. The kernel's slices capture alternating vertical motions (up/down). It seems that the network learns the sharp camera vibration that occurs when the runner's foot hits the ground. Fig. \ref{fig:kernel_driving} shows representative slices from the top two kernels associated with class \emph{Driving}. It is interesting to note that kernels learn `mixed' flow (inside the car and through the car's windows). While the learned flow vectors from inside the car are small and randomly oriented, the vectors from outside the car match those of forward motion.

\subsubsection{Evaluation by Transfer Learning}

One way of assessing a CNN's generalization capacity is by evaluating the transferability of its learned features to another dataset or task. We attempt this by first training our network on the $7$ new classes from the previous experiment (Sec. \ref{sec:egoseg_plus}). We then fix the weights for all layers in the network except the last, which is initialized randomly. We proceed by training only the last layer on a randomly chosen training subset of the original $7$ classes of \cite{us}. Concretely, we train a network with videos of the following $7$ classes: \emph{Boxing, Cooking, Skiing, Stair Climbing, Running, Horseback} and \emph{Sailing}. We then transfer the learned weights to a new network and retrain only its classification layer with videos from the following $7$ classes: \emph{Walking, Driving, Riding Bus, Standing, Sitting, Static} and \emph{Biking}.

We achieve $73.8\%$ classification accuracy (recall) on the complementary test subset in this setting. As expected, the results are inferior compared to when full network training is done end-to-end. While the accuracy is significantly less than the $89\%$ we report in Table~\ref{tb:egoseg_comparison}, it is still better than the current state-of-the-art. This implies that the learned features of the network are able to generalize to other egocentric activities.

To further explore this direction, we perform a slightly different experiment. This time, we use the weights learned from training on the $7$ new classes to initialize a new network, which we train end-to-end on the original $7$ classes. We use only half the training samples ($650$ instead of $1300$), run only $800$ iterations (instead of $3000$ iterations) and reach an accuracy of $82\%$. This implies again that the network indeed learns general motion features for egocentric videos. It also shows that given a pre-trained network, generalizing to new classes requires significantly less data and run-time. This is particularly important in egocentric vision due to the data scarcity problem.

\subsection{Discriminating Egocentric Videos}

\paragraph{Dataset:} We use the dataset-of-datasets provided by Tan et al. \cite{ego_or_not_cvprw14}, which contains more than $165$ hours of video (approximately $107$ hours of egocentric and $59$ hours of non-egocentric) covering a diverse set of activities, objects and locations. The egocentric datasets provided have considerable variation, making any attempt to characterize egocentric video as a whole very challenging. Some egocentric videos are shot indoors and with little whole-body movement, while others included walking from indoors to outdoors (and vice-versa). Furthermore, some videos were captured from cameras with wide-angle lenses. Non-egocentric videos were taken from $7$ different third-person video datasets containing sports, movie and surveillance footage.

\paragraph{Results:} We use the same experimental methodology as suggested by Tan et al. \cite{ego_or_not_cvprw14}. In the first experiment we divide each dataset in half, and use one half for training and the other for testing. This experiment is referred to as `Seen', since every dataset has some representation in the training set. The second experiment, referred to as `Unseen', is performed in an iterative manner. In each iteration, one dataset is left out from the training set and serves as a test set.  Our experiments indicate an almost perfect classification accuracy for the `Seen' experiments, while the accuracy for the `Unseen' experiments is $90.9\%$. See Table \ref{tb:ego_or_not_comparison} for a complete comparison. The high accuracy achieved by our network makes it practical for large scale video repositories such as YouTube, etc. This can also potentially automate collecting large amount of samples for new research topics in egocentric vision.

\begin{table}
    \centering
    \begin{tabular}{lrrrr}
        \toprule[1.5pt]
		& \multicolumn{2}{c}{\bf \tabletext Seen} & \multicolumn{2}{c}{\bf \tabletext Unseen}   \\
        \bf \tabletext 	& \bf \tabletext \cite{ego_or_not_cvprw14}      & \bf \tabletext Ours & \bf \tabletext \cite{ego_or_not_cvprw14}      & \bf \tabletext Ours\\ \midrule
 \bf Egocentric &  $71\%$  &  $\bf 99.1\%$  &  $ 62.7\%$  &  $\bf90\%$        \\
 \bf Non-Egocentric &  $99.3\%$  &  $\bf 99.4\%$  &  $ 67.1\%$  &   $\bf95.3\%$   \\
\midrule
 \bf Weighted Mean   		&   $    75.7\%$      &  $\bf 99.2\%$ &   $ 63.4\%$ 		  &   $\bf90.9\%$        \\
        \bottomrule[1.5pt] \\
    \end{tabular}
    \caption{Comparison with \cite{ego_or_not_cvprw14} for determining if a video is egocentric or not. Our method achieves high recall rates, making it practical for large scale search applications (e.g., automatic egocentric video collection).}
    \label{tb:ego_or_not_comparison}
\end{table}

\paragraph{Analysis:} We repeat the process of finding the kernels with the strongest affinity to each class and visualize them as optical flow fields. The kernels in this case were more difficult to decipher. We attribute this to the significant inner-class diversity, stemming from the unique properties of each dataset.

\section{Concluding Remarks}
\label{sec:concl}

A convolutional neural network architecture is proposed for long term activity recognition in egocentric videos. The input to the network is a sparse optical flow volume. The network improves the classification accuracy by $19\%$ compared to state-of-the-art for activity-based temporal segmentation. We extend the dataset by $7$ more classes and show that the proposed network is able to learn features for the new activities and achieve an overall recall rate of $86\%$ on the $14$ activities dataset. Another classification problem supported by our network is discriminating egocentric videos from non-egocentric videos. We achieve a near perfect accuracy of $99.2\%$ in this task. To understand what the network is learning, we visualize the learned kernel as flow fields and analyze affinities between kernels and classes.

\noindent\textbf{Acknowledgment:} This research was supported by Intel ICRI-CI and by Israel Science Foundation. Special thanks to Or Sharir for implementation of 3D convolution and pooling.
{\small
\bibliographystyle{ieee}
\bibliography{deepego}
}

\end{document}